\documentclass[runningheads]{llncs}
\usepackage[T1]{fontenc}
\usepackage{graphicx,verbatim}
\usepackage{graphicx}
\usepackage{cite}
\usepackage{amsmath,amssymb,amsfonts}
\usepackage{algorithmic}
\usepackage{graphicx}
\usepackage{textcomp}
\usepackage{adjustbox}
\usepackage{multirow}
\usepackage{bbold}
\usepackage{siunitx}
\usepackage{hyperref}       
\usepackage{xcolor}
\usepackage{sidecap}
\hypersetup{
    colorlinks,
    linkcolor={blue!50!black},
    citecolor={blue!50!black},
    urlcolor={blue!50!black}
}
\usepackage{amsmath, bm}
\usepackage{hyperref}

\usepackage{amssymb}
\usepackage{pifont}

\usepackage{floatrow}
\newfloatcommand{capbtabbox}{table}[][\FBwidth]

\usepackage{booktabs}
\usepackage{marvosym} 

\begin{document}
\title{Surgical Action Planning \\ with Large Language Models}
%
\author{Mengya Xu\inst{1}$^\star$ \and Zhongzhen Huang\inst{2,3}\thanks{Equal contribution.} \and Jie Zhang \inst{4}\and  \\
Xiaofan Zhang \inst{2,3} \and
Qi Dou \inst{1} \textsuperscript{\Letter}
}

\authorrunning{Mengya Xu et al.}
%

\institute{
The Chinese University of Hong Kong, Hong Kong SAR, China \and 
Shanghai Jiao Tong University, Shanghai, China  \and 
Shanghai AI Laboratory, Shanghai, China \and
Huazhong University of Science and Technology, Wuhan, China \\
}



\maketitle              


\begin{abstract}
In robot-assisted minimally invasive surgery, we introduce the Surgical Action Planning (SAP) task, which generates future action plans from visual inputs to address the absence of intraoperative predictive planning in current intelligent applications. SAP shows great potential for enhancing intraoperative guidance and automating procedures. 
However, it faces challenges such as understanding instrument-action relationships and tracking surgical progress. Large Language Models (LLMs) show promise in understanding surgical video content but remain underexplored for predictive decision-making in SAP, as they focus mainly on retrospective analysis. Challenges like data privacy, computational demands, and modality-specific constraints further highlight significant research gaps.
To tackle these challenges, we introduce \textbf{LLM-SAP}, a Large Language Models-based Surgical Action Planning framework that predicts future actions and generates text responses by interpreting natural language prompts of surgical goals. The text responses potentially support surgical education, intraoperative decision-making, procedure documentation, and skill analysis. LLM-SAP integrates two novel modules: the Near-History Focus Memory Module (NHF-MM) for modeling historical states and the prompts factory for action planning.
We evaluate LLM-SAP on our constructed CholecT50-SAP dataset using models like Qwen2.5 and Qwen2-VL, demonstrating its effectiveness in next-action prediction. Pre-trained LLMs are tested in a zero-shot setting, and supervised fine-tuning (SFT) with LoRA is implemented. Our experiments show that Qwen2.5-72B-SFT surpasses Qwen2.5-72B with a 19.3\% higher accuracy. 
\end{abstract}


\section{Introduction}  
In Robot-assisted Minimally Invasive Surgery (RMIS), current intelligent applications, such as surgical workflow recognition~\cite{ayobi2024pixel,jin2021temporal,cao2023intelligent}, instrument segmentation~\cite{ayobi2023matis, yu2024sam}, action recognition~\cite{kiyasseh2023vision, bai2024ossar}, and medical question-answering~\cite{bai2025surgical, seenivasan2023surgicalgpt}, primarily focus on retrospective analysis rather than future decision-making. While valuable for procedure analysis, they lack the ability to support intraoperative future decision-making, underscoring the need for our Surgical Action Planning (SAP) task, a forward-looking framework. 
Our SAP streamlines complex procedures by decomposing them into discrete actions, and generates long-horizon sequential action plans from visual inputs, enabling the achievement of user-defined objectives. For instance, natural language directives like \textit{``Conduct Calot’s triangle dissection, clipping and cutting of duct and vessel, and dissection of the gallbladder from the liver bed"} can be converted into a sequence of executable surgical actions, providing procedural guidance and demonstrating the potential for procedure automation.

Large language models (LLMs)~\cite{zhao2023survey} have shown impressive reasoning capabilities, enabling applications such as question answering~\cite{jiang2023reasoninglm}, machine translation~\cite{wang2023document}, and information extraction~\cite{jiao2023instruct}. Building on this foundation, recent advancements in Large Language Models (LLMs) and Vision Language Models (VLMs)  have demonstrated significant potential for understanding surgical video content, yet their application in future decision-making remains underexplored. For instance, while LLMs have been leveraged to refine and enrich surgical concepts through hierarchical knowledge augmentation \cite{yuan2024procedure}, these approaches primarily focus on retrospective analysis and comprehension rather than predictive planning. LLMs show great promise for the SPA task. Their ability to jointly align visual inputs with language prompts, parse complex scenes, ground language goals in visual contexts, and generate step-by-step plans makes them well-suited for SPA, as evidenced by their success in daily planning tasks like daily activities~\cite{chen2023egoplan,huang2023palm,kim2024palm}. Meanwhile, interpretable text analysis from LLMs can be utilized for surgical education prior to operations, providing decision support and assistance during critical intraoperative phases. It offers recommendations to surgeons, helps document and summarize surgical procedures, and serves as a valuable tool for postoperative skill analysis and improvement.

Additionally, fine-tuning LLMs on private data addresses data privacy concerns and enables task-specific customization, unlike zero-shot inference which relies on pre-trained language models and risks exposing sensitive information. However, fine-tuning large models is challenging due to their parameter sizes and resource demands. To address these limitations, efficient fine-tuning methods, such as supervised fine-tuning (SFT)~\cite{wei2021finetuned}, have been developed to reduce computational costs while maintaining performance.

The contributions of our work can be summarized as follows:
\begin{itemize} 
\item We introduced the Surgical Action Planning (SAP) task in RMIS, which generates future surgical action plans from visual inputs, focusing on forward-looking decision-making rather than retrospective analysis.
\item We developed the Large Language Models-based Surgical Action Planning framework (LLM-SAP) that predicts future actions by integrating two innovative modules: the Near-History Focus Memory module (NHFM) for modeling historical states, and the prompts factory for generating action plans.
\item We offered a flexible solution for customizing the zero-shot and fine-tuning capabilities of both LLMs and VLMs by thoughtfully designing the processing of visual observations, enabling the adaptation to both modalities.
\item We introduce Relaxed Accuracy (ReAcc), a novel evaluation metric that adopts a flexible approach to assess action forecasting. By flexibly considering action forecasting successful if they occur within the current or subsequent one step, ReAcc accounts for the dynamic and adaptive nature of surgical actions. 
\item We evaluate the effectiveness of the LLM-SAP framework on our constructed CholecT50-SAP dataset, derived from the CholecT50 dataset~\cite{nwoye2022rendezvous}, for surgical action planning. The evaluation involves testing with state-of-the-art (SOTA) LLM and VLM models, including Qwen2.5 and Qwen2-VL, under both zero-shot and fine-tuning experimental settings.


\end{itemize}

\section{Methods}
\subsection{Surgical action planning (SAP) task definition}
In SAP, the model generates an action plan $\mathcal{A} = \{a_1, \dots, a_t\}$ by leveraging two key inputs: a visual history $\mathcal{H}$ and a user-defined goal $G$, aiming to transition the current state to the desired goal within a planning horizon of $T$ steps. The visual history $\mathcal{H}$, represented as a sequence of video clips $\{v_1, \dots, v_t\}$, captures the progression toward the goal over time. Meanwhile, the goal $G$ is expressed as a natural language description, such as \textit{``Provide analysis and the next action for laparoscopic cholecystectomy."} Each action $a_t$ in the plan corresponds to a categorical label within a set of $C$ possible actions.

\begin{figure}[h]
\centering 
\includegraphics[width=1\linewidth]{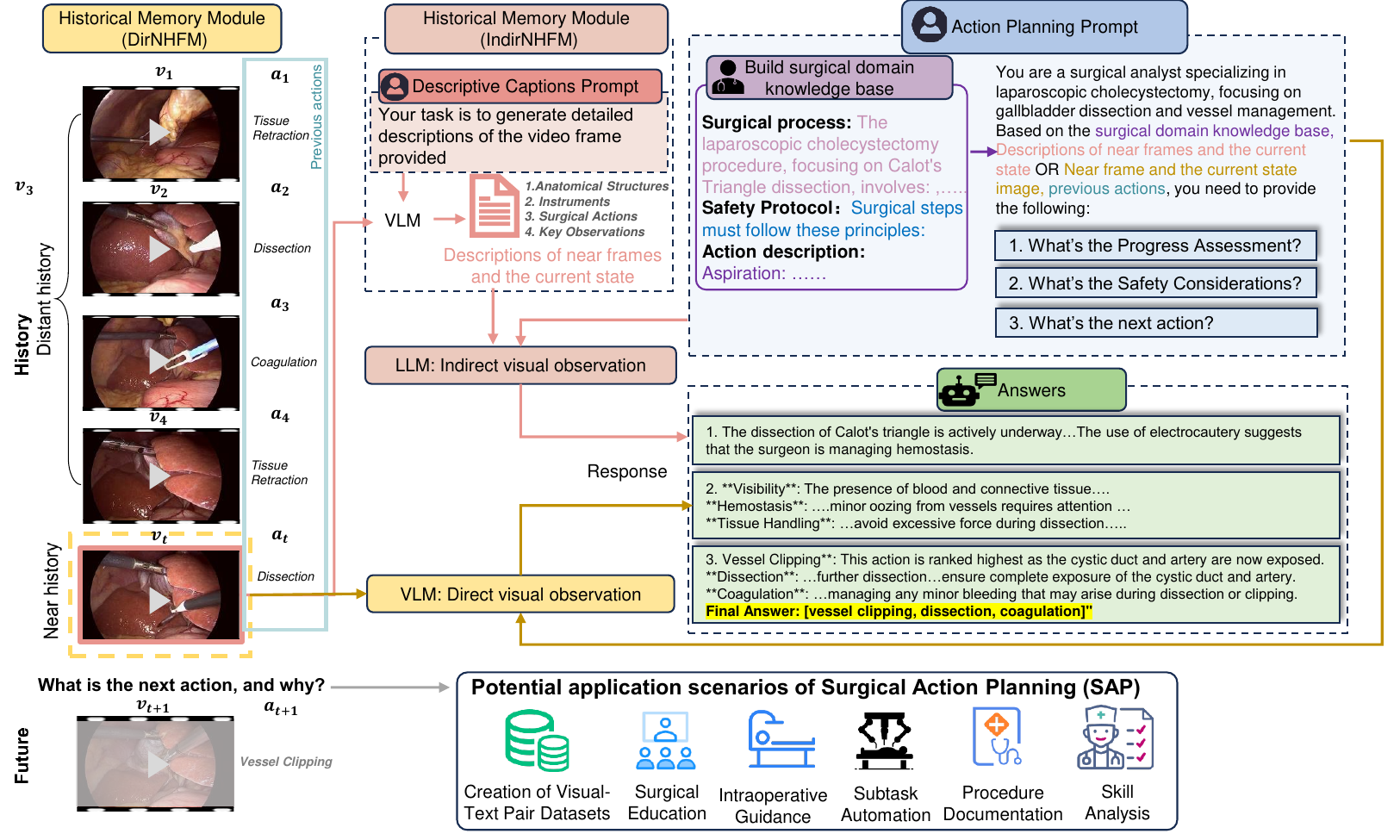} 
\caption{The architecture of our LLM-SAP. It is developed to predict the next surgical action and form long-horizon action chains by breaking procedures into a fixed set of actions, powered by advanced LLMs. LLM-SAP has two versions: a text-based LLM planning model that utilizes the text descriptions of the visual history in IndirNHFM, and a VLM planning model that uses visual history directly in DirNHFM, both guided by the action planning prompt. The data flow of these two versions is indicated by the pink and yellow lines, respectively. Given the surgical domain knowledge base and action planning prompts like \textit{``What's the next action?"}, LLMs analyze the history memory state to understand the surgical progress and generate structured responses, including progress assessment, safety considerations, and future action recommendations.}
\label{fig_MLLM_planning}  
\end{figure}

\subsection{Our LLM-SAP}
By breaking down complex surgical procedures into a fixed closed set of actions, we develop the LLM-SAP to predict the next action and link these actions into long-horizon action chains, powered by advanced LLMs (see Fig.~\ref{fig_MLLM_planning}). 
LLM-SAP has versions: one processes text descriptions of visual history, while the other directly analyzes visual history. Both are guided by action-planning prompts. By leveraging domain knowledge and prompts, the system generates a response in a structured format. The following sections will provide further details.

\subsubsection{Near-history focus memory module (NHFM)}
Previous studies on history understanding for VLM-based planners generally rely on a combination of historical action labels and their associated frames to understand the historical state. $\mathcal{H}_t =\mathrm{VLM}(\{\langle f_i,a_i\rangle\mid i= 1,\ldots,t\},\text{Prompt})$. However, incorporating lengthy action histories can overwhelm the planning process, often resulting in reduced performance \cite{chen2023egoplan, xie2024revealing}. In surgical settings, the next action is determined by the current tissue state and the procedure's dynamic progression, rather than adhering to a rigid sequence of past actions. To address this limitation, we introduce an enhanced history state understanding approach to build the Historical Memory Module (HMM) that emphasizes a compact summary of past actions for the distant history while focusing on detailed information from the near history.

\textbf{Direct visual observation for creating NHFM (DirNHFM):} ${a_i}$ for time steps 1 to $t-1$ encapsulates a compact representation of distant historical actions (action labels only), while $\langle v_t, a_t \rangle$ captures the detailed near history through the current video frame $v_t$ and its associated action label $a_t$. Additionally, \textit{Prompt} refers to the overarching prompt supplied to the VLM. It can be formulated as Equation~\ref{Eq. 1}. The details of the \textit{Prompt} design will be introduced in the next section.
\begin{equation}
\mathcal{A}_\text{DirNHFM} =\mathrm{VLM}(\{a_i\mid i=1,\ldots,t-1\},\langle v_t,a_t\rangle,\textit{Prompt})
\label{Eq. 1}
\end{equation}

\textbf{Indirect visual observation for creating NHFM (IndirNHFM):} To address the limitation of LLMs that do not support visual frames as input, we propose an indirect approach leveraging visual observations for HMM creation. Specifically, we use the VLM to process visual frames and generate descriptive captions. These captions serve as text-based input for LLMs unable to process visual data directly, enabling our framework to adapt to such LLMs (see Equation~\ref{Eq. 2} and~\ref{Eq. 3}).  
\begin{equation}
C_t = \mathrm{VLM}(v_t,\textit{DCPrompts})
\label{Eq. 2}
\end{equation}

\begin{equation}
\mathcal{A}_\text{IndirNHFM} =\mathrm{LLM}(\{a_i\mid i=1,\ldots,t-1\}, \langle C_t,a_t\rangle ,\textit{Prompt})
\label{Eq. 3}
\end{equation}

\subsubsection{Prompts factory for generating action plans}

Fig.~\ref{fig_MLLM_planning}) shows the prompts factory, including
(1) \textbf{Descriptive captions prompts (DCPrompts)} To generate descriptive captions using the VLM , we employ the following prompt: \textit{``You are a professional surgical analysis assistant specializing in laparoscopic cholecystectomy. Your task is to generate detailed descriptions of the video frame provided, focusing on anatomical structures, tool manipulation, key surgical steps, and environmental features.''}
(2) \textbf{Action planning prompts (APPrompts)}: Firstly, \textbf{surgical domain knowledge base} is build based on \textit{surgical process}, \textit{safety protocol}, and \textit{action description}. Next, the APPrompts include \textit{``Based on the provided Surgical Domain Knowledge Base, Descriptions of near frames and the current state (IndirNHFM) OR Video frames provided (DirNHFM), Previous actions, Last action, you need to provide the following: \textbf{progress assessment}, \textbf{safety considerations}, \textbf{ready-to-execute actions}: Provide three actions. For each action, provide a rationale explaining why it is ranked in that order.''}

\subsubsection{Zero-shot and supervised fine-tuning}
Building on our carefully designed historical memory module, we begin by evaluating the performance of open-source models in surgical action planning. Given that even current advanced models struggle with such tasks, our goal is to enhance the performance of open-source models in this domain. Based on prior efforts, we adopt a distillation-based approach to generate high-quality data from stronger models. In particular, we leverage GPT-4o, which has demonstrated exceptional performance, to generate fine-tuning data. In total, we obtain 118 samples for fine-tuning data (63 for IndirNHFM and 55 for DirNHFM). 

\subsection{Implementation details}
We used Llama-Factory~\cite{zheng2024llamafactory} to fine-tune the large language models (LLM and VLM) with LoRA~\cite{hu2022lora}. we conducted training over $50$ epochs. The fine-tuning of all LLMs was performed on $8$ NVIDIA A800 GPUs, using a learning rate of $1e-4$ and a batch size of $8$.

\section{Experiments and results} 
\subsubsection{Dataset}
\begin{figure}[h]
\centering 
\includegraphics[width=1\linewidth]{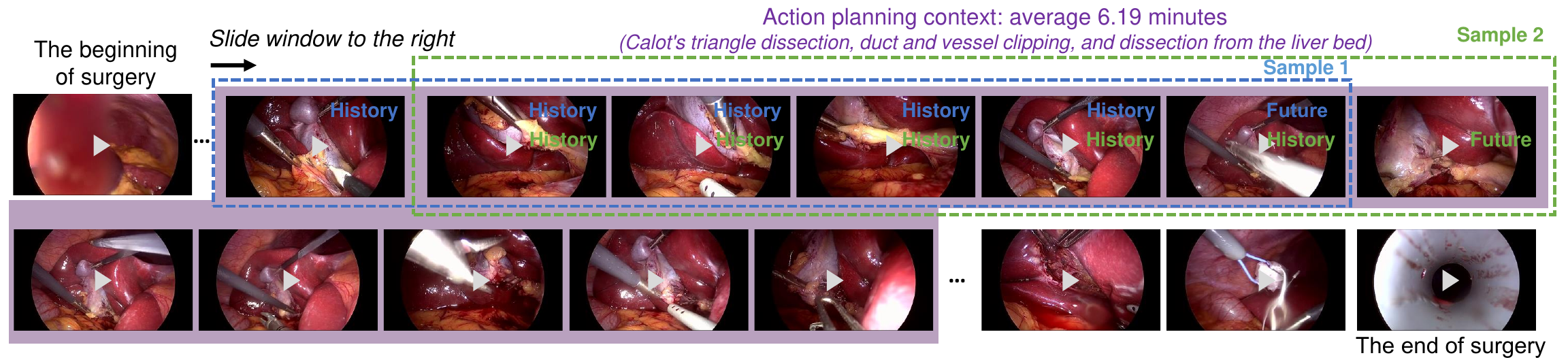} 
\caption{Example of the CholecT50-SAP dataset we constructed.}
\label{fig_dataset}  
\end{figure}

Our constructed CholecT50-SAP dataset is derived from the CholecT50 dataset~\cite{nwoye2022rendezvous}, which consists of 50 cholecystectomy surgical procedures. We grouped consecutive frames sharing the same action into action clips in CholecT50-SAP (see Fig.~\ref{fig_dataset}). The action labels include \{Aspiration, Coagulation, Dissection, Tissue Retraction, Vessel Clipping\}. We focus on the video segments encompassing \textit{``Calot's triangle dissection, duct and vessel clipping, and dissection from the liver bed"} as the context for implementing action planning. The $50$ chosen video segments have an average duration of $6.19$ minutes each. Including $5$ historical action clips, these $50$ video segments from cholecystectomy procedures comprise a total of $225$ samples. The dataset was split into training ($35$ videos, $168$ samples) and testing ($15$ videos, $57$ samples) sets, following~\cite{nwoye2023cholectriplet2022}.


\subsubsection{Metrics}
We report the sample and video-level accuracy under standard and relaxed conditions.

\textbf{Sample-level accuracy (SLAcc)} The standard SLAcc requires the predicted next action $\hat{a}_i$ to exactly match the ground truth action $a_i$ at each time step. On the other hand, Top-2 and Top-3 SLAcc evaluate the proportion of samples in which the true action label $a_i$ for the future video clip is included among the top two or three predictions, respectively. \textbf{Video-level accuracy (VLAcc)}
We report VLAcc to evaluate the model's performance on individual surgical patients. This is calculated by averaging the mean of SLAcc across all surgical patients. \textbf{Relaxed accuracy (ReAcc)}
Our proposed ReAcc adopts a more flexible evaluation approach, motivated by the need to account for the dynamic nature of surgical actions. It considers future action recommendations successful if the model's suggested actions $\hat{a}_i$ occur within the current step or the subsequent one-step $\{a_{i}, a_{i+1}\}$ ($\hat{a}_i \in \{a_{i}, a_{i+1}\}$), aligning with the variability and adaptability required in surgical procedures.

\subsubsection{Action planning results}
Building on an LLM-based planning framework, we employ multiple state-of-the-art (SOTA) LLM (Qwen2.5) and VLM (Qwen2-VL) to predict the probable next action. In addition to the zero-shot experiments, we also conduct the supervised fine-tuning (SFT) approach~\cite{wei2021finetuned} (see Table~\ref{tab:main}).

\begin{table}[htbp]
\scalebox{0.7}{
  \centering
  \caption{Comparison of performance across different methods on the CholecT50-SAP dataset, with the best results highlighted in bold.}
    \begin{tabular}{cccccccc|cccccc}
    \toprule
    \multirow{3}[2]{*}{Models} & \multirow{3}[2]{*}{HMM} & \multicolumn{6}{c|}{Standard Condition}        & \multicolumn{6}{c}{Relaxed Condition} \\
    \cmidrule{3-14}
          &       & \multicolumn{3}{c|}{SLAcc} & \multicolumn{3}{c|}{VLAcc} & \multicolumn{3}{c|}{Re SLAcc} & \multicolumn{3}{c}{Re VLAcc} \\
              \cmidrule{3-14}
          &       & Top1  & Top2  & Top3  & Top1  & Top2  & Top3  & Top1  & Top2  & Top3  & Top1  & Top2  & Top3 \\

    \midrule
    \multicolumn{14}{c}{\textit{\textbf{Zero-Shot}}} \\
    \midrule
    Qwen2.5-32B & \multirow{2}[1]{*}{IndirNHFM}      & 45.61 & 63.16 & 66.67 & \textbf{53.42} & 62.48 & 64.65 & 67.44 & 95.35 & \textbf{97.67} & 78.97 & \textbf{97.38} & 98.33 \\
    Qwen2.5-72B &       & 45.61 & 59.65 & 66.67 & \textbf{53.42} & 59.31 & 64.65 & 67.44 & 83.72 & \textbf{97.67} & 78.97 & 89.21 & \textbf{99.05} \\
\cmidrule{2-14}  
    Qwen2-VL & \multirow{1}[1]{*}{DirNHFM}      & 40.35 & 57.89 & 68.42 & 48.37 & 56.37 & 66.32 & \textbf{72.09} & 88.37 & 90.70 & \textbf{82.48} & 93.45 & 95.24 \\
    \midrule
    \multicolumn{14}{c}{\textit{\textbf{Supervised Fine-Tuning}}} \\
    \midrule
    Qwen2.5-72B-\texttt{SFT} & \multirow{2}[2]{*}{IndirNHFM} & 45.61 & \textbf{78.95} & 80.70 & \textbf{53.42} & 80.05 & 82.27 & 67.44 & 93.02 & \textbf{97.67} & 78.97 & 93.33 & 98.33 \\
    Qwen2.5-32B-\texttt{SFT} &       & 45.61 & \textbf{78.95} & \textbf{85.96} & \textbf{53.42} & \textbf{83.05} & \textbf{87.89} & 67.44 & \textbf{95.35} & \textbf{97.67} & 78.97 & 96.67 & 98.33 \\
    \cmidrule{2-14} 
    Qwen2-VL-72B-\texttt{SFT} & \multirow{1}[1]{*}{DirNHFM}      & \textbf{47.37} & 54.39 & 68.42 & \textbf{54.31} & 58.55 & 65.98 & 60.47 & 74.42 & 90.70 & 67.30 & 83.25 & 94.76 \\
    \bottomrule
    \end{tabular}%
  \label{tab:main}%
  }
\end{table}%

\begin{figure}[h!]
\centering 
\includegraphics[width=1\linewidth]{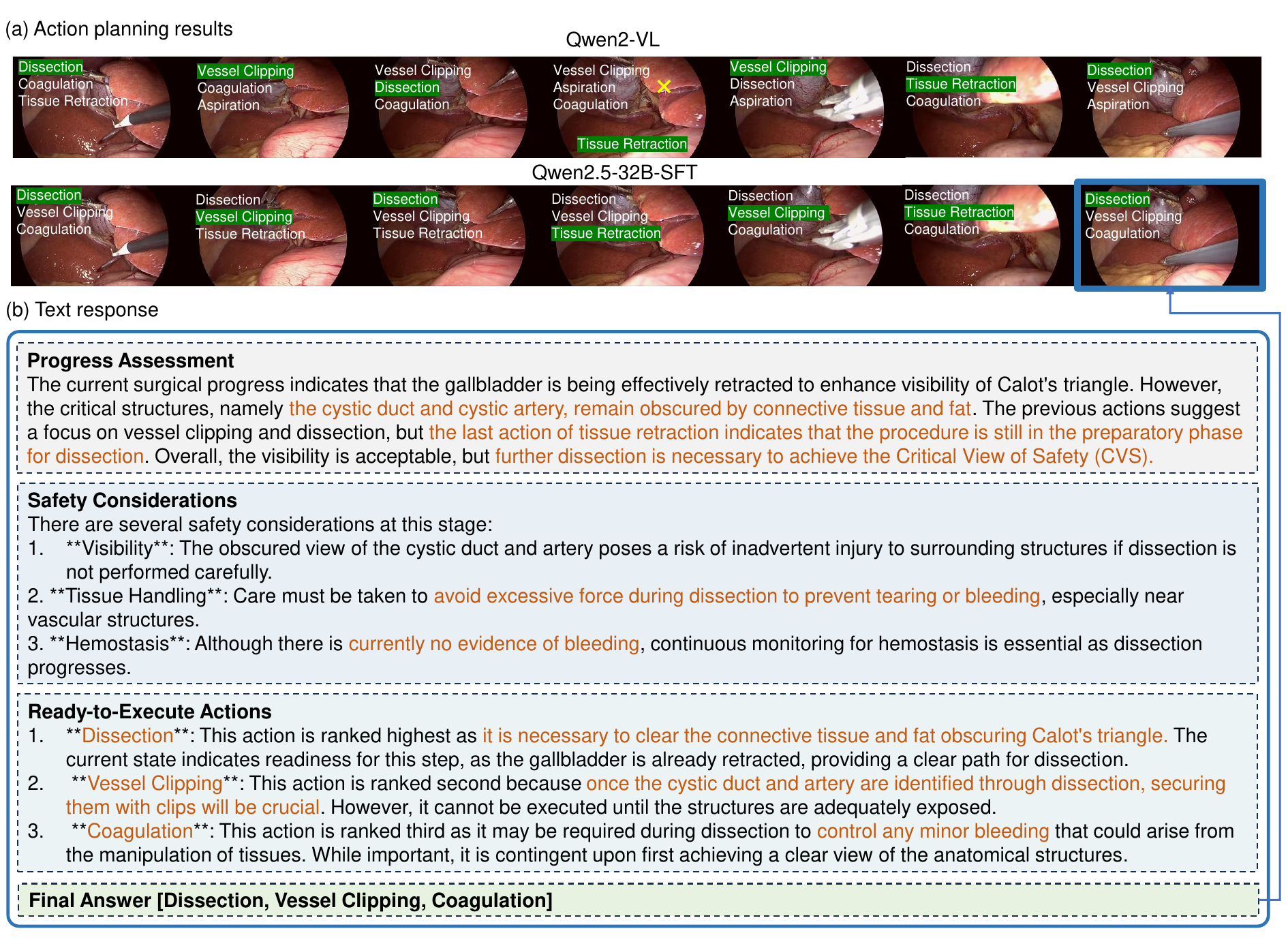} 
\caption{(a) Action planning results visualization. White text displays the planning results, showing the Top 3 future action predictions in order. ${ \times } $ indicates incorrect prediction. The text with a green background color indicates the ground truth actions. (b) Example of the text response used to derive the action planning answer.}
\label{fig_planning_results}  
\end{figure}

The analysis of the experimental results can be summarized as follows: (1) \textbf{IndirNHFM vs. DirNHFM:} In zero-shot experiments, Qwen2.5-72B achieves a standard top-1 SLAcc of 45.61\% and VLAcc of 53.42\% with IndirNHFM, outperforming Qwen2-VL with DirNHFM by 5.26\% (45.61\% vs. 40.35\%) in SLAcc and 5.05\% (53.42\% vs. 48.37\%) in VLAcc. This indicates that indirect visual observation better captures contextual nuances, potentially due to richer textual representations. (2) \textbf{Zero-Shot vs. SFT:} Qwen2.5-72B-SFT outperforms Qwen2.5-72B, achieving 19.3\% higher standard top-2 SLAcc (78.95\% vs. 59.65\%), 20.74\% higher standard top-2 VLAcc (80.05\% vs. 59.31\%), 9.3\% higher relaxed top-2 SLAcc (93.02\% vs. 83.72\%), and 4.12\% higher relaxed top-2 VLAcc (93.33\% vs. 89.21\%). This demonstrates that SFT significantly boosts model performance for SAP by leveraging task-specific data. (3) \textbf{Qwen2.5-32B-SFT \textit{(best model)} vs others:} Although Qwen2-VL-72B-SFT achieves the best standard top-1 SLAcc of 47.37 and VLAcc of 54.31, Qwen2.5-32B-SFT demonstrates strong overall performance across all metrics, with 44.3\% higher standard top-2 SLAcc (78.95\% vs. 54.39\%), 41.8\% higher standard top-2 VLAcc (83.05\% vs. 58.55\%), 28.1\% higher relaxed top-2 SLAcc (95.35\% vs. 74.42\%), and 16.2\% higher relaxed top-2 VLAcc (96.67\% vs. 83.25\%).

Fig.~\ref{fig_planning_results}(a) presents the action planning results for different LLMs and Fig.~\ref{fig_planning_results}(b) shows the interpretable reasoning processes for a single sample.

\begin{figure}[h!]
\centering 
\includegraphics[width=0.8\linewidth]{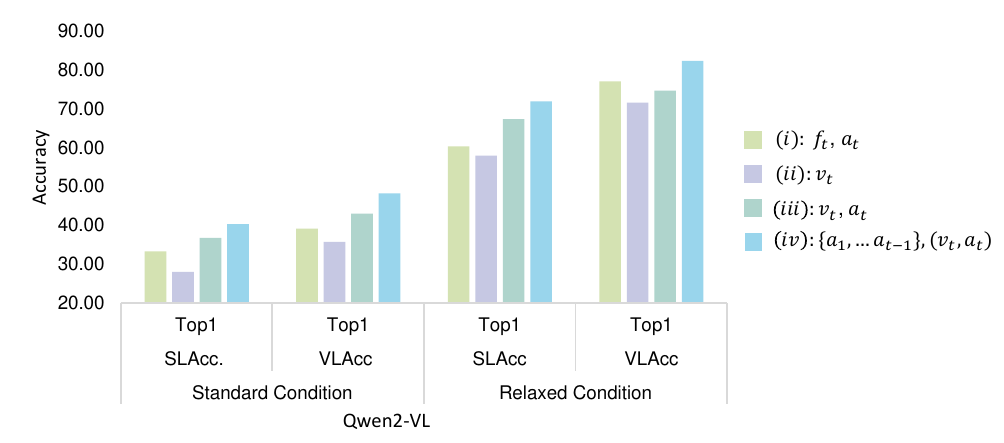} 
\caption{Ablation experiments on HMM creation based on Qwen2-VL.}
\label{fig_ablation}  
\end{figure}

\textbf{Ablation experiments on HMM creation}
We conducted ablation studies on the Historical Memory Module (HMM), as shown in Fig.~\ref{fig_ablation}.

(i) \textbf{with an image from near history, along with the last previous action}: $\{f_t,a_t\}$;
(ii) \textbf{with only near history}: $C_t$ or $v_t$
(iii) \textbf{with near history and its associated action}: $\langle v_t,a_t\rangle$;
(iv) \textbf{with previous action labels and near history}: $\{a_1,\ldots,a_{t-1}\}, \langle v_t,a_t\rangle $.
The comparison between settings (ii) and (iv) highlights the importance of previous action labels in modeling historical states. Among the four HMM creation methods, setting (iv) demonstrates the best results. Therefore, we adopt setting (iv) as the default method for creating the HMM.

\section{Conclusion}
We introduce the Surgical Action Planning (SAP) task in computer-assisted surgery, generating future action plans from visual inputs with a focus on forward-looking decision-making. This task has immense potential to enhance intraoperative guidance and procedural automation. 
Key challenges such as complex instrument-action relationships, temporal dependencies, progress tracking, and data privacy concerns have hindered progress in the SAP task. To address these challenges, we propose the LLM-SAP framework leveraging large language models to predict actions and provide interpretable responses by integrating our proposed NHF-MM and prompts factory. Evaluated on our constructed CholecT50-SAP dataset using state-of-the-art models such as Qwen2.5 and QwenVL, LLM-SAP demonstrated effectiveness in recommending future actions while effectively addressing data privacy through SFT. 
\textbf{Future work:} To enhance LLM-SAP, we will leverage reasoning-based LLMs to use text responses from earlier steps as input for subsequent steps, enabling prior-conditioned and continuous planning while highlighting differences in structured outputs. Additionally, we aim to expand the framework to support more procedures and robotic integration.



\bibliography{mybib}{}
\bibliographystyle{splncs04}
\end{document}